\documentclass[letterpaper]{article} 
\usepackage{aaai2026}  
\nocopyright
\usepackage{times}  
\usepackage{helvet}  
\usepackage{courier}  
\usepackage[hyphens]{url}  
\usepackage{graphicx} 
\usepackage{natbib}  
\usepackage{caption} 
\usepackage{algorithm}
\usepackage{algorithmic}
\usepackage{booktabs}
\usepackage{diagbox}
\usepackage{amsfonts}
\usepackage{amsmath}
\usepackage{etoolbox} 
\usepackage{newfloat}
\usepackage{listings}
\DeclareCaptionStyle{ruled}{labelfont=normalfont,labelsep=colon,strut=off} 
\floatstyle{ruled}
\newfloat{listing}{tb}{lst}{}
\floatname{listing}{Listing}
\title{Causal Reward Adjustment: Mitigating Reward Hacking in External Reasoning via Backdoor Correction}
\author{
    Ruike Song\textsuperscript{\rm 2,\rm 3}\equalcontrib,
    Zeen Song\textsuperscript{\rm 1,\rm 2}\equalcontrib,
    Huijie Guo\textsuperscript{\rm 1,\rm 2},
    Wenwen Qiang\textsuperscript{\rm 1,\rm 2}\thanks{Corresponding author},
}
\affiliations{
    \textsuperscript{\rm 1}University of Chinese Academy of Sciences,\\
    \textsuperscript{\rm 2}Institute of Software Chinese Academy of Sciences,\\

    \textsuperscript{\rm 3}Nankai University College of Software\\

}

\usepackage{bibentry}
\pubnote{Preprint. Under Review.}

\begin{document}

\maketitle

\begin{abstract}
External reasoning systems combine language models with process reward models (PRMs) to select high-quality reasoning paths for complex tasks such as mathematical problem solving. However, these systems are prone to reward hacking, where high-scoring but logically incorrect paths are assigned high scores by the PRMs, leading to incorrect answers.
From a causal inference perspective, we attribute this phenomenon primarily to the presence of confounding semantic features.
To address it, we propose Causal Reward Adjustment (CRA), a method that mitigates reward hacking by estimating the true reward of a reasoning path.
CRA trains sparse autoencoders on the PRM’s internal activations to recover interpretable features, then corrects confounding by using backdoor adjustment. Experiments on math solving datasets demonstrate that CRA mitigates reward hacking and improves final accuracy, without modifying the policy model or retraining PRM.
\end{abstract}


\section{Introduction}
Large language models (LLMs) have been widely recognized as a central focus of contemporary artificial intelligence research \cite{Shao_2024,zhao2023survey}.
One prominent direction in LLM research is external reasoning, which relies on external reward signals such as Process Reward Models (PRMs) to guide the reasoning process \cite{lightmanLetsVerifyStep2023, snellScalingLLMTestTime2024}. This approach has demonstrated strong performance on complex tasks such as mathematical problem solving \cite{weiChainofthoughtPromptingElicits2022, lightmanLetsVerifyStep2023, uesato2022solving, liu2025can}. However, external reasoning systems are vulnerable to a critical flaw known as \textbf{reward hacking}, where reward models assign higher scores to incorrect reasoning steps than to correct ones \cite{skalse2022defining}. As a result, the system might select high-score steps that are logically wrong, which ultimately reduces the accuracy of the final output.

To better understand the mechanism underlying the reward hacking phenomenon, we adopt a causal inference perspective and formulate the problem using a Structural Causal Model (SCM). In our formulation, the input reasoning path is denoted as $X$, the score as $Y$, and semantic features unrelated to correctness as $Z$. These features may include stylistic elements, step length, or frequently used expressions. Ideally, logically correct reasoning paths ($X$) should causally lead to higher reward scores ($Y$), represented by the direct path $X \rightarrow Y$. However, in observed training data, due to annotation biases or inherent preferences of annotators, reasoning paths that contain certain semantic patterns ($Z$) frequently receive higher labeled rewards. As a result, the presence of these semantic features ($Z$) creates a spurious correlation between reasoning paths ($X$) and reward scores ($Y$), represented by a backdoor path $X \leftarrow Z \rightarrow Y$. Here, the semantic feature $Z$ acts as a confounder, leading to a confounding effect where the observed distribution $\mathbb{E}[Y \mid X]$ no longer accurately represents the true causal relationship $X \rightarrow Y$. Consequently, the PRM trained on such biased observational data learns to assign rewards based not only on logical correctness ($X \rightarrow Y$) but also on the presence of semantic features ($Z \rightarrow Y$), causing the reward hacking phenomenon. This confounding effect can be addressed using a backdoor adjustment. Specifically, it involves evaluating how a given reasoning path $X$ would be scored under different values of $Z$. These scores are then averaged, weighted by how frequently each semantic feature occurs in the data. This neutralizes the influence of the confounder and recovers the true causal effect.

Inspired by the above causal analysis, we propose a causally grounded method called \textbf{Causal Reward Adjustment (CRA)}. CRA consists of three steps: (1) extracting interpretable features from the reward model, (2) identifying confounding features associated with reward hacking, and (3) implementing backdoor adjustment using the identified features to mitigate the reward hacking problem. In the first step, we train a sparse autoencoder (SAE) on the hidden representations of PRM. The SAE learns to encode each internal representation into a sparse latent vector, where each dimension corresponds to a distinct semantic feature. This design ensures that the latent space is interpretable and suitable for identifying the reward hacking feature. In the second step, we detect which latent feature is the reward hacking feature. We compute the activation distribution of each feature over two groups of reasoning steps, those labeled as reward hacking and those considered normal, and apply two-sample $t$-tests to quantify their statistical separation. Features with both statistically significant differences and sufficiently high activation are selected as potential reward hacking features. In the third step, we perform backdoor adjustment by marginalizing over the identified features. Specifically, we simulate the reward scores that would be assigned to a given reasoning path under different values of the confounding features, and compute a weighted average using their empirical frequencies in the training data. This procedure removes the influence of spurious correlations and yields an unbiased estimate of the true causal effect of the reasoning path on the reward, thereby directly addressing the reward hacking problem.

To evaluate the effectiveness of CRA, we conduct experiments on two mathematical reasoning benchmarks: GSM8K and MATH. Results show that CRA significantly mitigates reward hacking and improves the reasoning correctness. Furthermore, ablation studies show that intervening on features identified by our CRA method effectively suppresses reward hacking, while random interventions have little impact.  These findings validate both our causal analysis and the effectiveness of the proposed intervention strategy.
Our main contributions are summarized as follows:
\begin{itemize}
\item We provide a causal explanation for reward hacking, showing that semantic confounders induce spurious correlations between reasoning steps and rewards. As a result, the reward model may assign high scores to incorrect reasoning steps simply because they contain preferred stylistic patterns.
\item We propose CRA, a causally grounded three-step method that (i) extracts interpretable features via sparse autoencoders, (ii) identifies features responsible for reward hacking, and (iii) performs backdoor adjustment to eliminate their spurious influence, thereby mitigating the reward hacking effect.
\item Experiments on GSM8K and MATH show that CRA reduces reward hacking and improves reasoning performance, demonstrating the effectiveness of our approach.
\end{itemize}

\section{Related Works}

\textbf{Inference Methods in Large Language Models.} 
Recent advances in LLMs have led to the development of external reasoning methods, which enhance complex problem solving by decoupling inference-time reasoning from model parameters. These approaches construct multiple candidate reasoning paths using frozen LLMs, and rely on sampling- or search-based mechanisms guided by an external verifier to select high-quality solutions~\cite{lightmanLetsVerifyStep2023,wuInferenceScalingLaws2024,snellScalingLLMTestTime2024a,yaoTreeThoughtsDeliberate2023,selAlgorithmThoughtsEnhancing2024,bestaGraphThoughtsSolving2024,zhangAutomaticChainThought2022,brownLargeLanguageMonkeys2024,liu2025can}. 
Compared to internal approaches that improve reasoning via reinforcement learning~\cite{madaanSelfRefineIterativeRefinement2023,saunders2022self,deepseek-aiDeepSeekR1IncentivizingReasoning2025,shaoDeepSeekMathPushingLimits2024}, 
external methods offer flexibility and modularity, enabling dynamic inference-time optimization without retraining. This paradigm has shown strong performance on math-reasoning benchmarks such as MATH~\cite{hendrycksmath2021}, and GSM8K~\cite{cobbe2021gsm8k}.

\textbf{Reward Hacking.}
Although PRMs have shown strong capabilities in guiding LLMs during external reasoning, they are also prone to misuse. In some cases, a model may produce reasoning steps that receive high scores from the PRM, yet clearly deviate from correct logic or human judgment. This “high score but incorrect” behavior is a typical case of reward hacking. The term refers to situations where a model finds unintended shortcuts in the reward function.
This issue is proposed in reinforcement learning~\cite{weng2024rewardhack,skalse2022defining,pan2024feedback,liu2024rrm}.
As a result, models often “game the system.” They maximize reward signals in ways that violate the designer’s original goals. This phenomenon is captured by Goodhart’s Law\cite{goodhart2015goodhart}: “when a measure becomes a target, it ceases to be a good measure.” Amodei et al.\cite{amodei2016concreteproblemsaisafety} identified reward hacking as a core safety concern, emphasizing the risks of models exploiting flawed objectives. To address this problem, researchers have explored several approaches, including information-theoretic regularization~\cite{miao2024inform}, as well as intent-inference frameworks such as Inverse Reward Design (IRD) and Cooperative Inverse Reinforcement Learning (CIRL)~\cite{hadfieldmenell2020inverserewarddesign}, which aim to recover human preferences from observed behavior.

\section{Problem Formulation}

In this section, we formalize the reward hacking problem in external reasoning tasks. We first describe the external reasoning framework, where policy models generate reasoning paths scored by reward models, then define reward hacking as assigning a high score to an incorrect reasoning path.

\subsection{Solving Reasoning Problem with External Reward}
Reasoning tasks typically involve problems and their ground-truth answers drawn from a distribution $(x, y^*) \sim D$. The primary goal of external reasoning systems is to maximize the expected accuracy of solving these problems. To achieve this, external reasoning relies on PRMs to identify and select high-quality reasoning paths.
An external reasoning system comprises three main components:
(i) A policy model $\pi_\theta: \mathcal{X} \times R^{(t-1)} \rightarrow \Delta(S)$. Given an input problem $x \in \mathcal{X}$ and the sequence of previously generated reasoning steps $r^{(t-1)} = (s_1, s_2, \ldots, s_{t-1})$, the policy model generates a distribution over possible next reasoning steps $s_t \in S$. Here, $\pi_\theta$ is a language model parameterized by $\theta$, $\mathcal{X}$ represents the space of input problems, $S$ is the set of potential reasoning steps, and $\Delta(S)$ denotes the probability distribution over $S$.
(ii) A PRM $R_\phi: \mathcal{X} \times R^{(t)} \rightarrow \mathbb{R}$, which evaluates the quality of a reasoning trajectory up to the current step $r^{(t)} = (s_1, s_2, \ldots, s_t)$ and provides a numerical score. The PRM, parameterized by $\phi$, assigns higher scores to trajectories that better reflect accurate or desirable reasoning.
(iii) A search algorithm, such as beam search, which utilizes these reward scores to iteratively guide the construction of reasoning paths. 

The external reasoning process using beam search begins with an empty reasoning trajectory for a given problem $x$. At each time step $t$, the policy model $\pi_\theta$ proposes candidate next steps for each partial trajectory currently retained. These expanded trajectories are then scored by the reward model $R_\phi$. Next, the top-$K$ highest-scoring trajectories are preserved for further expansion, while lower-scoring alternatives are discarded. This iterative cycle of expansion and selective pruning continues until complete reasoning paths emerge. Finally, the output $(\hat{r}, \hat{y})$ is selected by:
\begin{equation}
(\hat{r}, \hat{y}) = \mathop{\arg\max}_{(r,y) \in \text{Complete}} R_\phi(x, r)
\end{equation}
where $\text{Complete}$ is the set of all fully formed reasoning paths terminating with a final answer.

The ultimate goal of external reasoning is defined as maximizing the expected accuracy over the distribution \(D\):
\begin{equation}
\mathbb{E}_{(x,y^*) \sim D}[\mathbb{I}[\hat{y} = y^*]]
\end{equation}
where \( \mathbb{I}[\cdot] \) is the indicator function representing correct predictions, and \( y^* \) denotes the ground-truth answer corresponding to input problem \( x \). In summary, achieving this goal critically depends on the PRM's ability to accurately score reasoning trajectories that lead to correct answers.

\begin{figure}[tb]  
\centering
\includegraphics[width=1\linewidth]{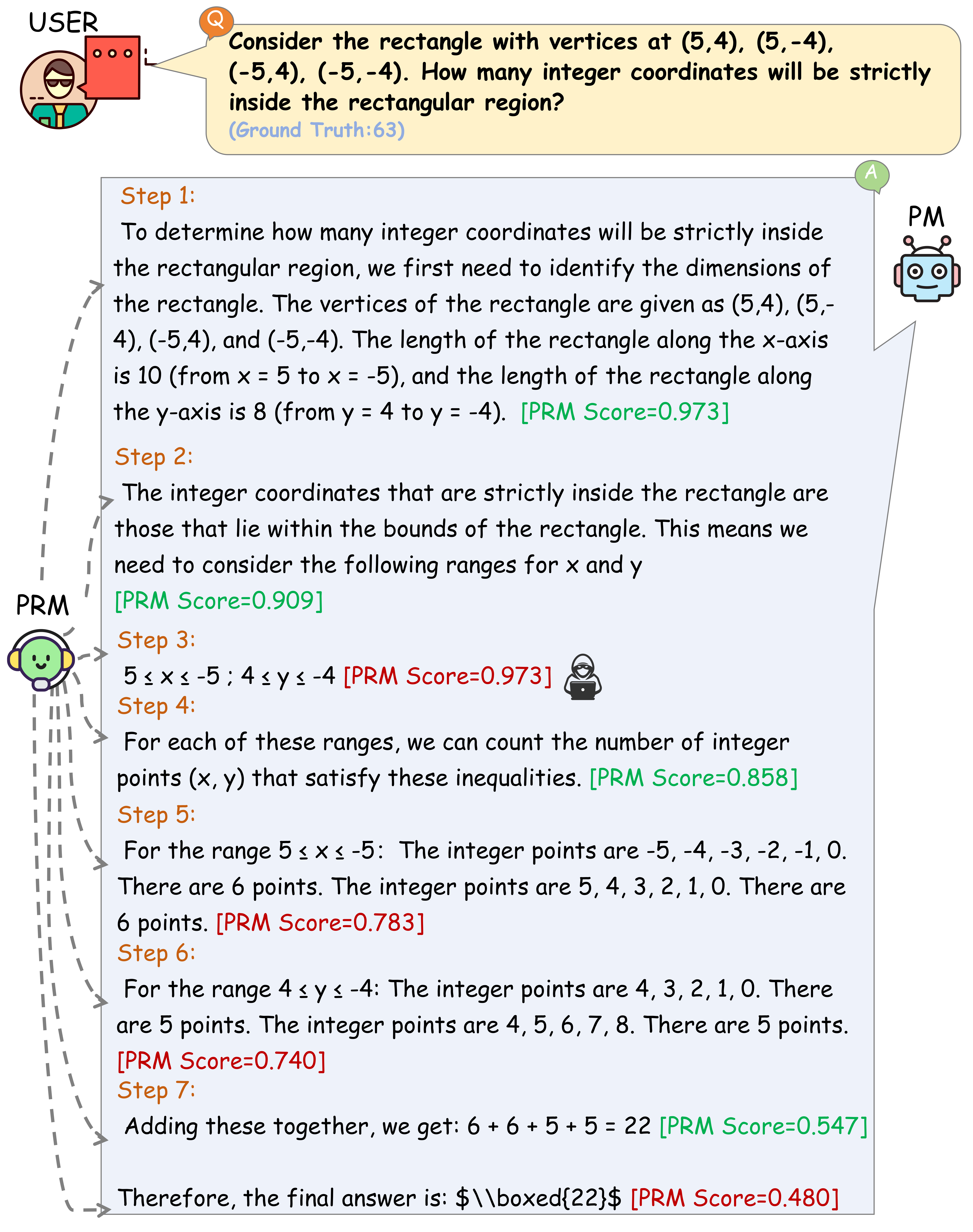}
\caption{Example of reward hacking in mathematical reasoning. PRM scores are shown in brackets: green indicates correct steps, red indicates erroneous steps. The little hacker icon identifies reward hacking instances where logically flawed steps receive high scores.}
\label{figure:reward_hacking_example}
\end{figure}

\subsection{The Reward Hacking Problem}
External reasoning systems rely on PRMs to rank candidate reasoning paths. However, these models occasionally assign high scores to reasoning steps that are not mathematically correct. As a consequence, the system tends to favor incorrect paths with higher scores, rather than correct ones with lower scores. This phenomenon is known as reward hacking.

To illustrate the reward hacking phenomenon, we present a concrete example from mathematical reasoning in Figure~\ref{figure:reward_hacking_example} demonstrates an instance of reward hacking. Specifically, Step 3 introduces constraints ``$5 \leq x \leq -5$; $4 \leq y \leq -4$", which are mathematically impossible since no number can simultaneously satisfy being greater than 5 and less than -5. Despite this clear logical contradiction, the reward model assigns a high score of 0.973 to this step. This led to an incorrect final answer.

\section{Causal Analysis}

Duging external reasoning, we observed that PRM occasionally assign high scores even to reasoning steps that are logically incorrect. 
To better understand the underlying causes of this phenomenon, we adopt a causal inference perspective. Specifically, we identify that the issue stems from the confounding effects induced by specific semantic patterns during the reward evaluation process. Based on the analysis, we argue that the true reward of a reasoning step can be estimated through backdoor adjustment, motivating novel approaches to mitigating reward hacking.

\subsection{Structural Causal Model}
\begin{figure}[t]  
\centering
\includegraphics[width=0.45\linewidth]{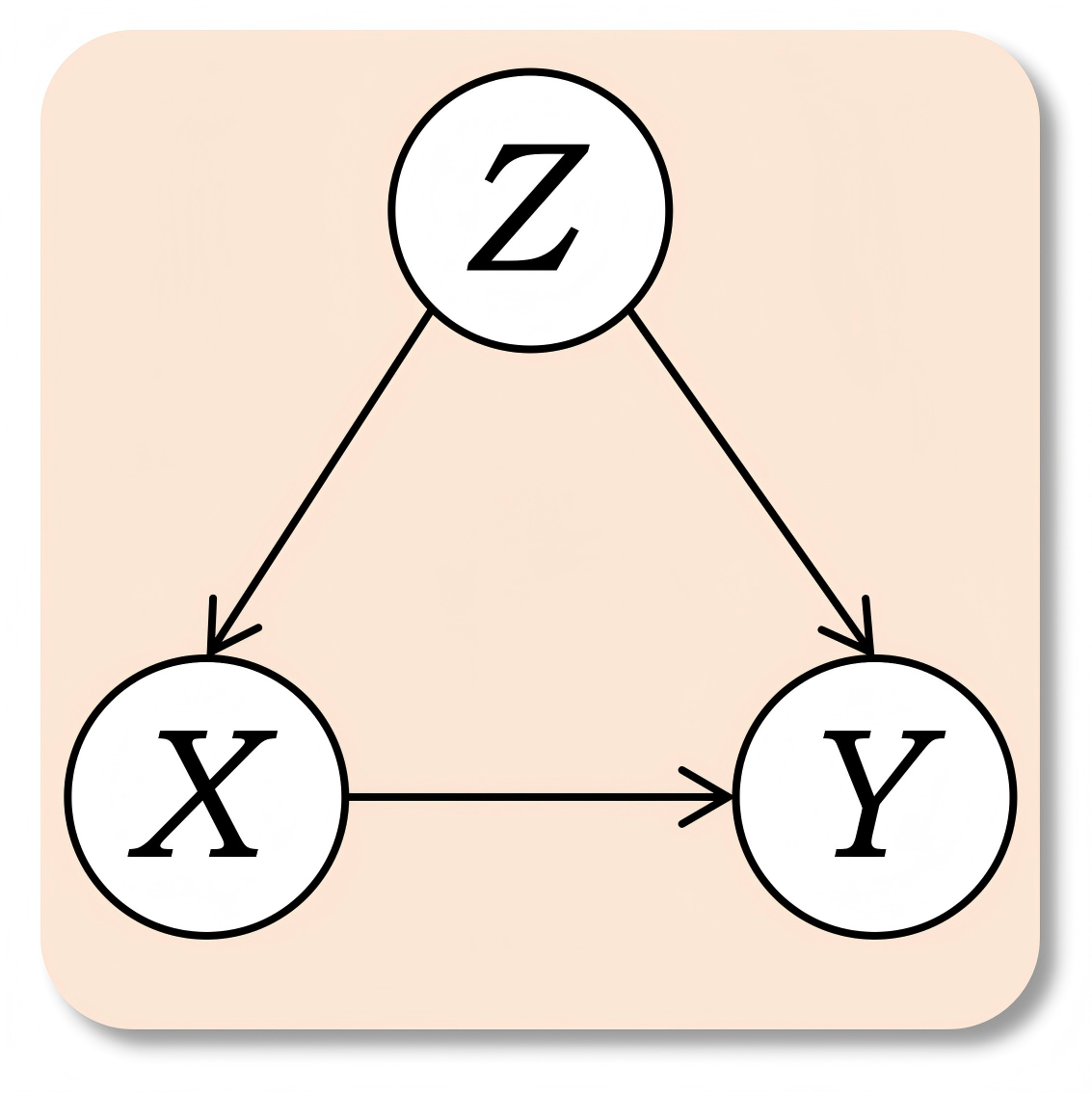}
\caption{The SCM showing the relationship between spurious patterns (Z), reasoning paths (X), and reward scores (Y).}
\label{figure:causal_diagram}
\end{figure}

We formalize the problem using a Structural Causal Model (SCM) to describe causal relationships between observational variables \cite{pearl2016causal}, as illustrated in Figure~\ref{figure:causal_diagram}. Let variable $X$ represent the sequence of reasoning steps, $Y$ the correctness score that reflects the logical validity of the reasoning, and $Z$ the semantic features unrelated to correctness, such as stylistic elements, step length, or frequently used expressions.

In the SCM, the directed edge $X \rightarrow Y$ signifies that coherent and logically correct reasoning paths $X$ should causally increase the reward score $Y$. The edge $Z \rightarrow X$ means that semantic patterns $Z$, such as step length, stylistic phrases, or frequently used templates, often appear in the generated reasoning steps. These features influence how the model constructs $X$. The edge $Z \rightarrow Y$ means that human annotators or automated annotation processes tend to assign higher scores to reasoning steps that contain such semantic patterns, regardless of their logical correctness~\cite{zhangLessonsDevelopingProcess2025}. Together, $X \leftarrow Z \rightarrow Y$ forms a backdoor path, introducing a spurious correlation between $X$ and $Y$.

\subsection{The Confounding Effect}
The structure of the SCM described above implies that, in observational data, the correctness score $Y$ is influenced not only by the logical validity of the reasoning path $X$ ($X \rightarrow Y$), but also by the presence of certain semantic features $Z$ ($X \gets Z \rightarrow Y$). Specifically, $Z$ acts as a confounder, simultaneously affecting the generation of reasoning steps and the labels assigned during annotation.

In practice, the PRM is trained to predict the probability that $Y=1$ by approximating the conditional expectation $\mathbb{E}[Y \mid X]$. In the presence of confounding, the conditional expectation learned by the PRM can be decomposed as:
\begin{equation}
    \mathbb{E}[Y \mid X] = \sum_{z\in\{0,1\}} \mathbb{E}[Y \mid X, Z = z] \, P(Z = z \mid X).
\end{equation}
When annotation biases yield $P(Y = 1 \mid Z = 1) \gg P(Y = 1 \mid Z = 0)$, the term $\mathbb{E}[Y \mid X = x, Z = 1]$ is close to one while $\mathbb{E}[Y \mid X = x, Z = 0]$ is near zero. Therefore, the conditional expectation simplifies to:
\begin{equation}
\label{eq:confounding}
    \mathbb{E}[Y \mid X = x] \approx P(Z = 1 \mid X = x).
\end{equation}
Equation \ref{eq:confounding} shows that the PRM learns to predict whether the confounding feature $Z$ is present in the reasoning path, rather than assessing the true logical validity of $X$, directly leading to the reward hacking phenomenon.

\subsection{Motivation}
Given that the PRM’s prediction $\mathbb{E}[Y \mid X = x]$ is biased by the confounding influence of $Z$, we seek to recover the true causal effect of the reasoning path on reward by removing this bias. Specifically, the backdoor adjustment formula states that the true causal effect of $X$ on $Y$, represented as $\mathbb{E}[Y\mid \mathrm{do}(X = x)]$, can be estimated by marginalizing over the confounder $Z$ according to its marginal distribution:
\begin{equation}
\label{eq:backdoor}
    \mathbb{E}[Y \mid \mathrm{do}(X)] = \sum_{z} \mathbb{E}[Y \mid X, Z = z] \; P(Z = z).
\end{equation}
This means that, for any given reasoning path $X = x$, the estimated reward is computed as if the confounding feature $Z$ could take either value, regardless of whether $x$ actually contains $Z$ or not. In practice, it is common to assume a balanced distribution, i.e., $P(Z = 1) = P(Z = 0) = 0.5$, so that neither presence nor absence of the confounder dominates the estimated score. As a consequence, the adjusted reward is independent of spurious semantic patterns, thereby effectively eliminating reward hacking. This motivates using a backdoor adjustment to mitigate reward hacking.

\begin{figure*}[tb]  
\centering
\includegraphics[width=1.0\linewidth]{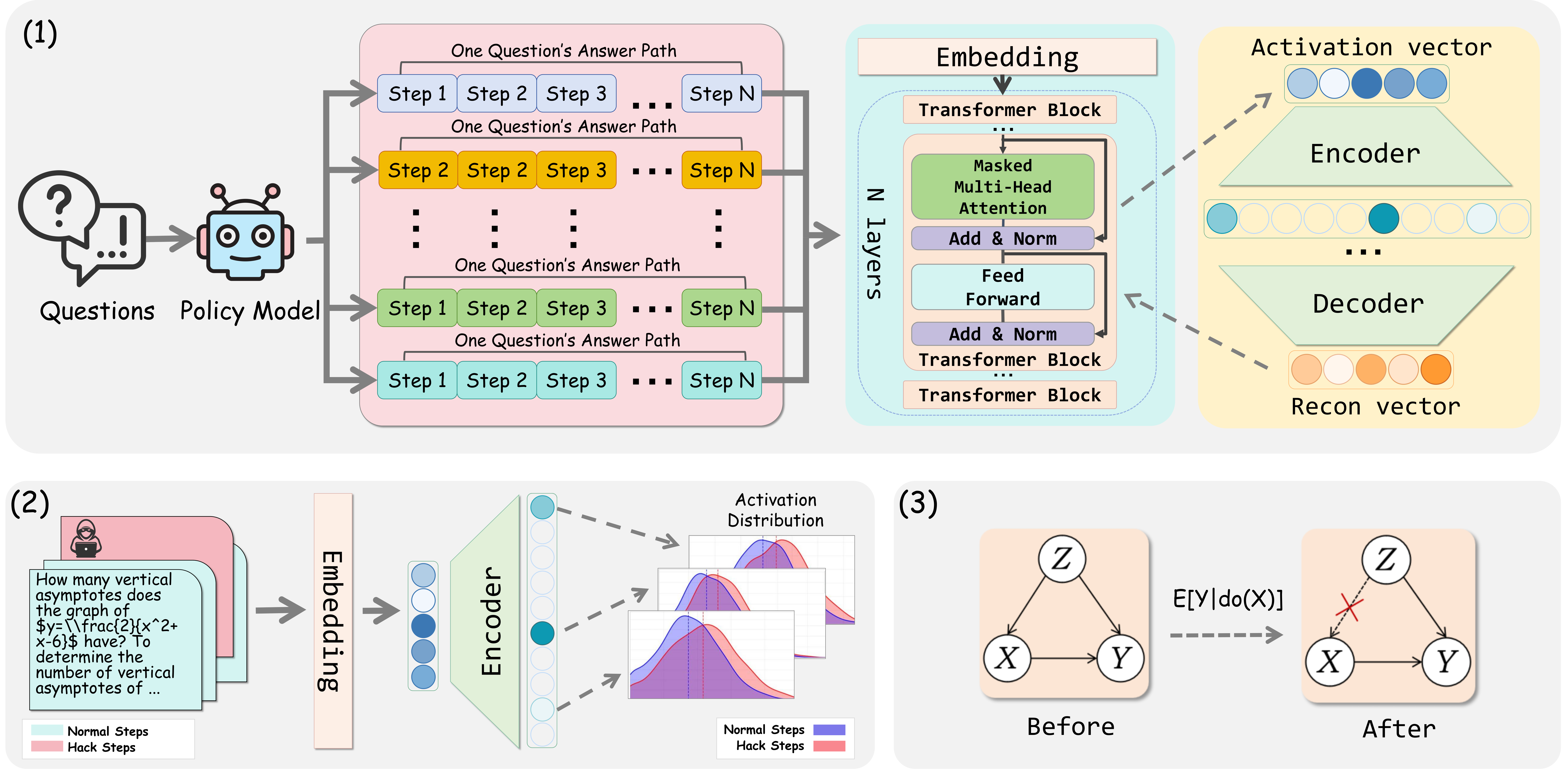}
\caption{\textbf{(1) Training Sparse Autoencoders on Reward Models.} Given some input problems, the policy model generates reasoning steps. These steps are then passed through selected transformer blocks in the reward model to extract step-level activation vectors. The activations are used to train SAEs. \textbf{(2) Identifying Reward Hacking Semantics.} The reasoning steps are manually labeled as normal (blue) or reward hacking (red with a hacker icon). These labeled samples are then encoded using the SAE trained in (a), and the activation distributions of individual features are obtained. Those with significant distribution differences are regarded as reward hacking features. \textbf{(3) Implementing Backdoor Adjustment.} Based on the reward hacking features identified in the previous step, the influence of these features during the scoring process can be cut off through backdoor regulation, thereby addressing the reward hacking issue.}
\label{figure:reward_hacking_example2}
\end{figure*}

\section{Methodology}
Grounded in our causal analysis, we propose Causal Reward Adjustment (CRA), a method designed to address reward hacking through backdoor adjustment.  CRA consists of three steps: (1) Training SAEs on reward models to decompose internal activations into interpretable semantic features, (2) Identifying reward hacking semantics by statistically analyzing which features discriminate between reward hacking and normal reasoning steps, and (3) Implement backdoor adjustment using the identified features to mitigate the reward hacking problem and improve system accuracy. The overall framework is illustrated in Figure \ref{figure:reward_hacking_example2}.

\begin{figure*}[t]  
\centering
\includegraphics[width=.95\linewidth]{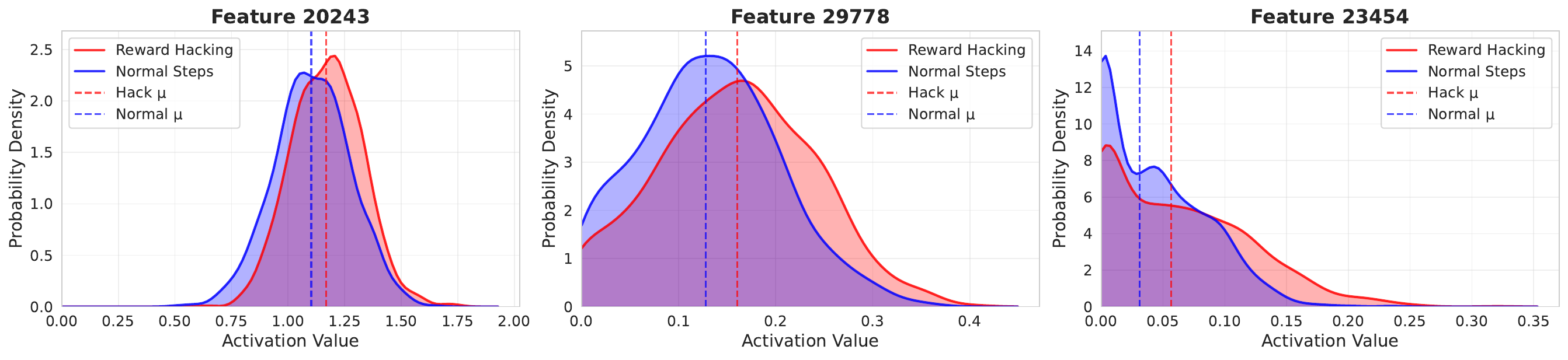}
\caption{Examples of features with high $t$-statistics showing systematic differences between reward hacking (red) and normal (blue) reasoning steps. Dashed vertical lines indicate mean activation values for each condition.}

\label{figure:feature_distributions}
\end{figure*}

\subsection{Training Sparse Autoencoders on Reward Models}
To identify interpretable semantic features within the reward model that may contribute to reward hacking, we train a set of SAEs to decompose the internal activations of the model
into sparse, disentangled components.

\textbf{Token-Level Activation Collection.}
To collect the training data for training SAEs, we first construct a corpus of reasoning paths by prompting multiple instruction-tuned LLMs (Qwen2.5-0.5B-Instruct, Qwen-2.5-math-7B-Instruct, LLaMA-3.1-8B-Instruct) to perform step-by-step reasoning on problems from training sets of GSM8K and MATH. Each resulting path consists of a sequence of intermediate reasoning steps, which are then processed token-by-token by the reward model $R_\phi$ to extract internal activations. And for each reasoning path, we tokenize the input and extract the hidden representation $h_{l,k} \in \mathbb{R}^d$ for each token position $k$ from each Transformer block $l \in \{1, ..., L\}$. These token-level activations will serve as input data for layer-specific autoencoders.

\textbf{Layer-Wise SAE Architecture.}
We train one SAE for each Transformer block layer in the reward model. 
At each layer $l$, the SAE encoder maps the hidden activation at token position $k$ to a sparse feature vector:
\begin{equation}
    z_{l,k} = \mathrm{ReLU}(W_e^{(l)} h_{l,k} + b_e^{(l)}),
\end{equation}
and the decoder reconstructs the original activation:
\begin{equation}
    \hat{h}_{l,k} = W_d^{(l)} z_{l,k}.
\end{equation}
Here, $h_{l,k} \in \mathbb{R}^d$ denotes the hidden representation of token $k$ at layer $l$ in the reward model, and $z_{l,k} \in \mathbb{R}^m$ is the corresponding sparse latent vector. The encoder consists of a weight matrix $W_e^{(l)} \in \mathbb{R}^{m \times d}$ and a bias vector $b_e^{(l)} \in \mathbb{R}^m$, while the decoder $W_d^{(l)} \in \mathbb{R}^{d \times m}$ uses the transpose of $W_e^{(l)}$ to reconstruct the input, i.e., $W_d^{(l)}=W_e^{(l)\top}$ Here, $d$ is the hidden size of the reward model, and $m$ is the number of latent features in the sparse representation. We set $m = 8d$ to promote overcomplete and disentanglement representations.

We train the SAE at each layer $l$ by minimizing the following empirical loss over the training set $\mathcal{H}_l$ of token-level activations:
\begin{equation}
\mathcal{L}^{(l)} = \frac{1}{|\mathcal{H}_l|} \sum_{h_{l,k} \in \mathcal{H}_l} \left[ \frac{1}{d} \| h_{l,k} - \hat{h}_{l,k} \|_2^2 + \alpha \| z_{l,k} \|_1 \right].
\end{equation}
Here, $\mathcal{H}_l$ denotes the set of hidden activations collected from layer $l$ across all tokens and reasoning paths in the training corpus, $\|\cdot\|_2$ denotes the Euclidean norm, and $\|\cdot\|_1$ denotes the $\ell_1$ norm. The first term ensures accurate reconstruction of the reward model's activations, and the second term enforces sparsity in the latent features. Besides those, the sparsity coefficient $\alpha$ is a hyperparameter.

\textbf{Feature Interpretation.}
After training, the sparse autoencoder provides interpretable semantic features. Specifically, each row vector $f_i^{(l)} \in \mathbb{R}^d$ of the decoder weight matrix $W_d^{(l)}$ represents a distinct semantic feature in layer $l$, effectively serving as a basis vector capturing specific patterns in the activation space.
To understand how these features work in practice, consider an input token sequence $T = [k_1, k_2, \dots, k_N]$. The hidden activation $h_{l, k_N} \in \mathbb{R}^d$ at the final token $k_N$ in layer $l$ can be decomposed as a linear combination of these learned semantic basis vectors:
\begin{equation}
    h_{l, k_N} \approx \sum_{i=1}^m z_{l,T}^{(i)} f_i^{(l)} = W_d^{(l)} z_{l,T},
\end{equation}
where $z_{l, T} \in \mathbb{R}^m$ is the sparse code computed by the encoder over the input sequence $T$. Each coefficient $z_{l,T}^{(i)}$ reflects the presence and strength of the $i$-th semantic feature in the representation of $T$:
if $z_{l,T}^{(i)} = 0$, the feature $f_i^{(l)}$ is inactive; if large, it strongly influences the output. Due to the enforced sparsity, only a few features are active per input, making their contribution interpretable. Empirically, these features align with symbolic computations, common logical substeps, or spurious patterns exploited by reward hacking \cite{bricken2023monosemanticity}. In the next section, we analyze these features to identify those correlated with reward hacking.

\subsection{Identifying Reward Hacking Semantics}
After obtaining SAEs, we can decompose the reward model activations into interpretable features with them. Next, we identify which specific features are responsible for reward hacking behavior. This process involves two key steps: (1) constructing a labeled dataset of reward hacking versus normal reasoning steps, (2) applying statistical analysis to identify discriminative features. Through this analysis, we can pinpoint the features that cause reward models to favor reward hacking steps.

\textbf{Dataset Construction for Feature Analysis.} To identify reward hacking features, we construct labeled examples by analyzing reasoning trajectories from the beam search process.
We isolate every intermediate reasoning step and assign binary labels based on two criteria: mathematical validity and reward score.
Specifically, we label $y_i = 1$ for steps that are mathematically incorrect yet receive high reward scores (reward hacking instances), and $y_i = 0$ for all other steps (normal instances).
This labeling process yields $N_1$ reward hacking steps and $N_0$ normal steps, providing sufficient data for statistical analysis.

\textbf{Statistical Feature Selection.} Building on our causal analysis, we aim to leverage SAEs to identify semantic features responsible for reward hacking. To achieve this, we propose a statistical screening approach: Each labeled reasoning step is first encoded into a sparse representation $z_i \in \mathbb{R}^m$ using the trained SAEs, where $m$ denotes the number of learned semantic features. We then identify reward hacking features by selecting those whose activation distribution differs significantly between reward hacking steps and normal reasoning steps. 

For each sparse dimension $j$, we compute the mean activations $\mu_{1,j}$ and $\mu_{0,j}$, and variances $\sigma^2_{1,j}$ and $\sigma^2_{0,j}$ within the two classes. We then compute the two-sample $t$-statistic:
\begin{equation}
t_j = \frac{\mu_{1,j} - \mu_{0,j}}{\sqrt{\sigma^2_{1,j}/n_1 + \sigma^2_{0,j}/n_0}}
\end{equation}
where $n_1$ and $n_0$ are the numbers of reward hacking and normal samples, respectively.

Features with high $|t_j|$ values represent the confounding variable $Z$ in our SCM, since they exhibit systematic differences between the two conditions, indicating their role in creating spurious correlations between reasoning paths and reward scores. We retain dimensions satisfying both statistical significance and activation thresholds:
\begin{equation}
|t_j| > \tau_t \text{ and } \max(\mu_{1,j}, \mu_{0,j}) > \tau_a
\end{equation}
where $\tau_t$ controls statistical significance and $\tau_a$ filters out low-activation noisy dimensions. We show in Figure~\ref{figure:feature_distributions} the probability distributions of activation values for representative features.
The vertical dashed lines indicate the mean activation values for each condition, showing systematic shifts between the two distributions. Having identified the observable confounding features $\mathcal{F}^* = {j : |t_j| > \tau_t \land \max(\mu_{1,j}, \mu_{0,j}) > \tau_a}$, we can implement causal intervention based on backdoor adjustment.

\begin{table*}[t]
\centering
\caption{Performance comparison using beam search (beam = 4) across different policy models. $\star$ represents our trained models.}
\label{tab:performance_comparison}
\resizebox{\textwidth}{!}{%
\begin{tabular}{lcccccc}
\toprule
\textbf{Policy Model} & \multicolumn{2}{c}{\rule{0pt}{2.3ex}\textbf{Qwen2.5-0.5B-Instruct}\rule[-1.0ex]{0pt}{0pt}} 
& \multicolumn{2}{c}{\rule{0pt}{2.3ex}\textbf{Qwen-2.5-math-7B-Instruct}\rule[-1.0ex]{0pt}{0pt}} 
& \multicolumn{2}{c}{\rule{0pt}{2.3ex}\textbf{Llama-3.2-3B-Instruct}\rule[-1.0ex]{0pt}{0pt}} \\
\cmidrule{1-7}
\diagbox[dir=NW]{\textbf{Reward Model}}{\textbf{Dataset} \quad } & \textbf{MATH} & \textbf{GSM8K} & \textbf{MATH} & \textbf{GSM8K} & \textbf{MATH} & \textbf{GSM8K} \\
\cmidrule{1-7}
\rule{0pt}{11pt}Math-Shepherd-PRM-7B & 40.1 & 55.1 & 77.0 & 96.8 & 48.3 & 78.1 \\[2pt]
Qwen2.5-Math-PRM-7B & 46.6 & 60.9 & 78.1 & 96.5 & 53.9 & 80.1 \\[2pt]
\midrule
$\star$ Math-Shepherd-PRM-7B + \textbf{CRA} & 43.7 & 58.0 & 80.3 & \textbf{97.1} & 51.7 & 80.7 \\[2pt]
$\star$ Qwen2.5-Math-PRM-7B + \textbf{CRA} & \textbf{48.6} & \textbf{62.3} & \textbf{80.6} & 97.0 & \textbf{56.4} & \textbf{82.1} \\
\bottomrule
\end{tabular}%
}
\end{table*}

\subsection{Implementing Backdoor Adjustment}
Once we identify which SAE latent corresponds to a reward-hacking semantic feature, we can perform backdoor adjustment based on this latent. This process consists of four steps:
(1) We collect the activation values of the reward-hacking feature across the dataset to estimate its prior distribution.
(2) For each reasoning step, we substitute the activation value of the reward-hacking feature with different possible values, reconstruct the hidden state, and obtain the corresponding conditional rewards from the PRM.
(3) The final CRA score, $\hat{R}_{\text{CRA}}$, is computed as the weighted average of all conditional rewards, using the estimated prior as weights.
(4) We then use $\hat{R}_{\text{CRA}}$ to replace the original PRM scores when selecting reasoning paths in external reasoning.

\textbf{Building Prior Distribution.}
According to Equation~\ref{eq:backdoor}, implementing backdoor adjustment requires estimating two components: the prior distribution $P(Z)$ and the conditional rewards $\mathbb{E}[Y \mid X, Z]$. To estimate the prior, we first collect the activation values of the reward-hacking feature across all reasoning steps. These activation values are then grouped into non-overlapping bins to construct an empirical distribution. For each bin $i$ with range $[z_i, z_{i+1}]$, the prior probability is calculated as
\begin{equation}
P(Z \in [z_i, z_{i+1}]) = \frac{n_i}{N},
\end{equation}
where $n_i$ is the number of steps within the bin and $N$ is the total number of steps. The midpoint of each bin is used as the intervention value $z$ in subsequent calculations.

\textbf{Compute Conditional Rewards.}
To obtain $\mathbb{E}[Y|X, Z = z]$, for each reasoning step $t$, we edit the SAE latent vector $z_t$ by replacing the activation value at the target dimension with $z$, while keeping all other dimensions unchanged:
\begin{equation}
\tilde{z}_t^{(j)} =
\begin{cases}
z, & \text{if } j \in \mathcal{F}^\star \\
z_t^{(j)}, & \text{otherwise}
\end{cases}
\end{equation}
We then decode $\tilde{z}_t$ to obtain the hidden state $\tilde{h}_t = W_d \tilde{z}_t$, and input it into the PRM. The resulting PRM output is taken as the conditional reward $\mathbb{E}[Y|X, Z = z]$.

\textbf{Perform Backdoor Adjustment.}
Finally, the adjusted reward is computed as the weighted average over all conditional rewards:
\begin{equation}
\hat{R}_{\text{CRA}}(x) = \sum_z \mathbb{E}[Y \mid X, Z = z] \cdot P(Z = z).
\end{equation}
By this construction, $\hat{R}_{\text{CRA}}(x)$ removes the PRM’s bias toward the semantic feature $Z$, thereby mitigating the reward hacking problem.

\textbf{System Integration and Evaluation.}
We integrate the adjusted reward into the external reasoning pipeline by replacing the original PRM score with the adjusted reward $\hat{R}_{\text{CRA}}$ at each reasoning step. During beam search, this adjusted score guides the selection of partial trajectories, which downgrades the rewards of incorrect steps even if they contain the semantic feature $Z$.

\section{Experiment And Result}
In this section, we evaluate the effectiveness of CRA by comparing external reasoning accuracy before and after CRA integration on mathematical benchmarks, assessing whether it improves the reasoning accuracy.

\subsection{Experimental Setup}

\textbf{Datasets.}
We evaluate our method on GSM8K\cite{cobbe2021gsm8k} and MATH\cite{hendrycksmath2021}. For SAE training, we construct a corpus of 18,000 reasoning trajectories using multiple policy models, comprising over 190,000 individual reasoning steps.

\textbf{Models and Architecture.}
We use Qwen2.5-0.5B-Instruct as the primary policy model, with additional evaluations on Qwen-2.5-math-7B-Instruct and Mistral-7B-Instruct. For reward models, we employ Math-Shepherd-Mistral-7B and Qwen2.5-Math-PRM-7B. We train layer-wise SAEs with sparse dimension $m = 8d$ to encourage interpretable feature discovery.

\textbf{Implementation Details.}
We implement our approach using PyTorch 2.0 on 8 Tesla V100 GPU with mixed-precision training. Beam search is configured with a beam size of 4. At each expansion step, 8 candidate steps are generated, and 8 complete reasoning paths are produced as final outputs. SAEs are trained using the Adam optimizer (lr=0.001, cosine annealing) for 50 epochs with a batch size of 2,048 and a sparsity coefficient $\alpha = 0.001$. For feature selection, we use thresholds $\tau_t = 4.0$ and $\tau_a = 0$ to identify discriminative features for intervention.

\subsection{Results}
The results are shown in Table~\ref{tab:performance_comparison}. From Table \ref{tab:performance_comparison}, we can observe that CRA consistently improves performance in all settings. On MATH, the average accuracy increases from 57.3\% to 60.2\%, with an average gain of 2.9 percentage points. On GSM8K, the accuracy improves from 77.9\% to 79.5\%, with an average gain of 1.6 percentage points. Specifically, in the combination of Qwen2.5-Math-PRM-7B and LLaMA-3.2-3B-Instruct as the policy model, GSM8K accuracy improves from 80.1\% to 82.1\%, indicating that CRA provides consistent benefits even in strong model configurations. These results confirm that CRA mitigates reward hacking and improves external reasoning reliability.

\subsection{Ablation Experiment}
\begin{figure}[t]  
\centering
\begin{minipage}{0.48\linewidth}
   \includegraphics[width=\linewidth]{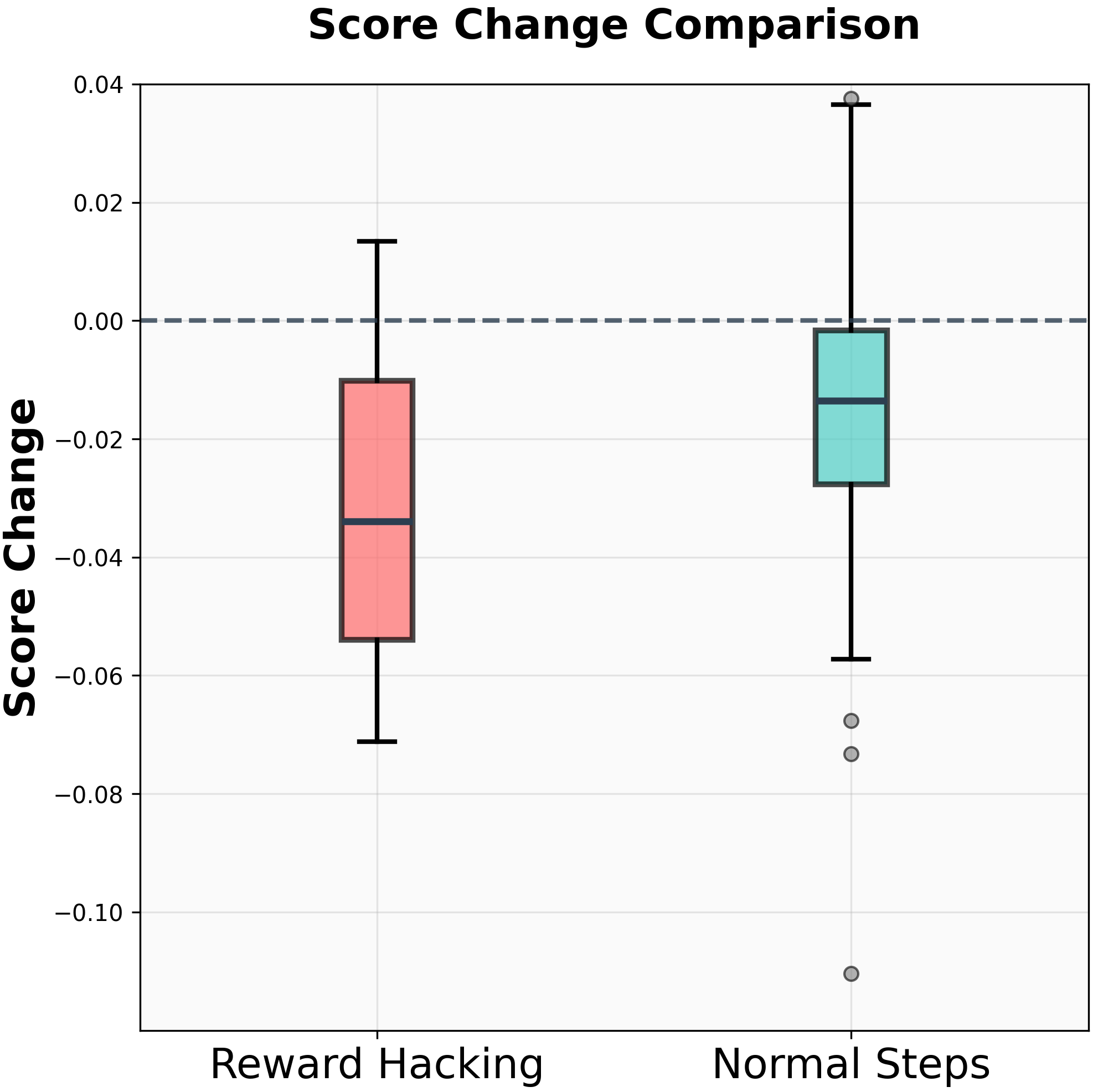}
   \caption*{(a) Causal intervention}
\end{minipage}
\hfill
\begin{minipage}{0.48\linewidth}
   \includegraphics[width=\linewidth]{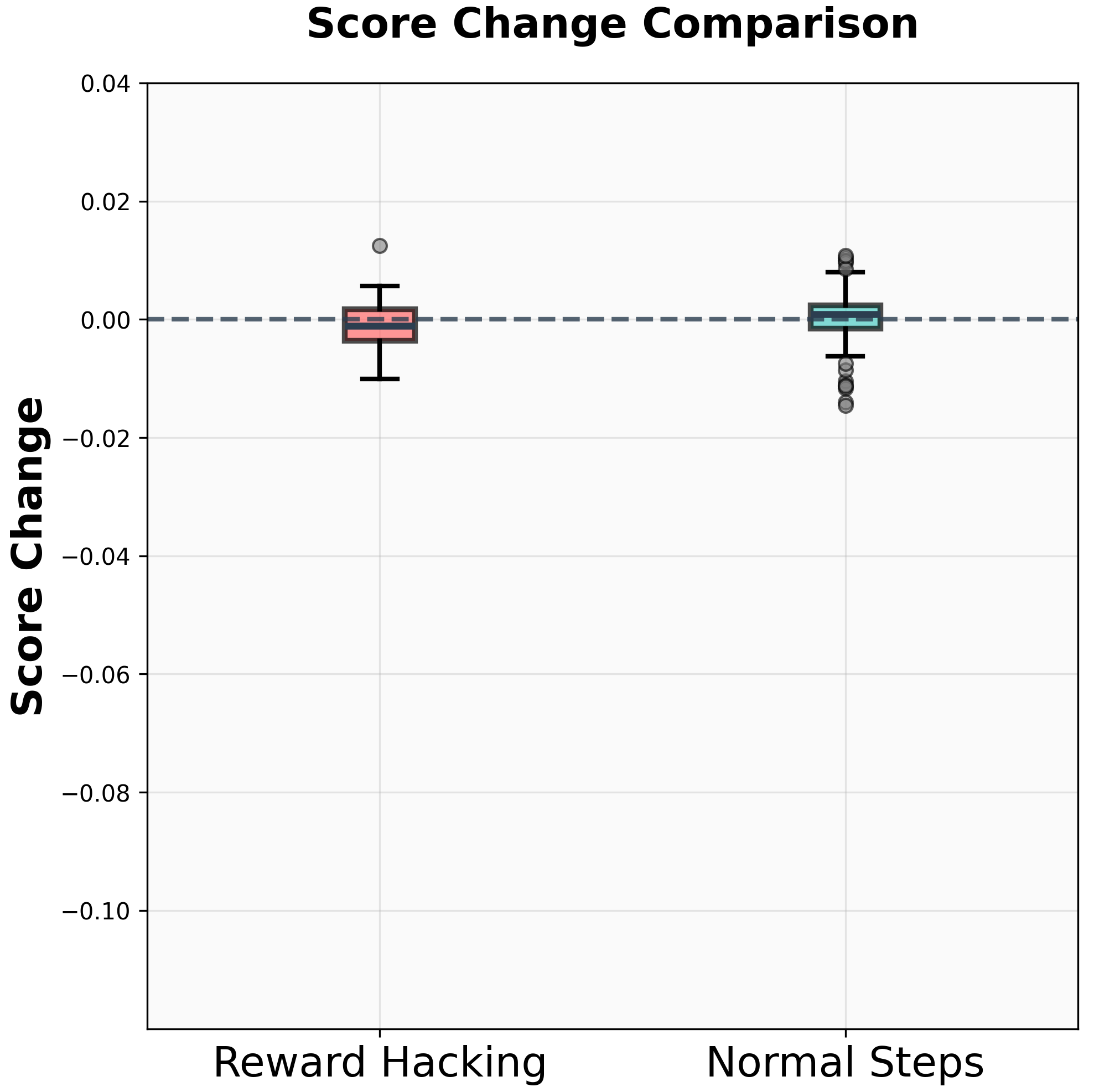}
   \caption*{(b) Random intervention}
\end{minipage}
\caption{Score change distributions after feature intervention. (a) Causal intervention effectively reduces scores for reward hacking instances while minimally affecting normal reasoning steps. (b) Random intervention shows minimal and non-discriminative effects on both step types.}
\label{figure:intervention_comparison}
\end{figure}
We conduct ablation studies to validate our CRA by comparing it against a random intervention baseline. Both methods share the same technical pipeline, differing only in feature selection: CRA intervenes on features identified by their statistical significance (highest $t$-statistics), while the random baseline selects features arbitrarily.

Experimental results (Figure~\ref{figure:intervention_comparison}) show a clear advantage for CRA. When intervening on selected features, CRA specifically lowers the scores of reward-hacking steps (average decrease $\approx -0.04$) without significantly impacting normal reasoning steps. Conversely, random feature interventions produce negligible and non-discriminative effects, with score changes tightly clustered around zero. These findings confirm that precise, causally-informed feature identification is critical for effectively mitigating reward hacking.
\section{Conclusion}
We analyze the reward hacking problem in external reasoning systems from a causal inference perspective. Our key finding is that reward hacking stems from confounding semantic features that simultaneously influence reasoning generation and reward evaluation, creating spurious high-scoring phenomena.
Inspired by this, we propose CRA, which uses SAEs to identify confounding features and applies backdoor adjustment to estimate debiased rewards. Experimental results validate our approach: CRA achieves significant improvements on both GSM8K and MATH datasets by reducing erroneous selections of flawed reasoning steps.
This work demonstrates the effectiveness of causal intervention for mitigating reward hacking without modifying policy models or retraining reward models.

\bibliography{main}

\section*{Appendix}
\label{sec:appendix}
The appendix is organized into several sections:

\begin{itemize}
    \item \textbf{Appendix Notations} provides details for all notations used in this paper.
    \item \textbf{Appendix Additional Related Work} Provide additional related works that are closely aligned with our proposed method.
    \item \textbf{Appendix Datasets} provides details for the datasets used in this paper.
    \item \textbf{Appendix LLMs} provides details for the LLMs used in this paper.
    \item \textbf{Appendix Additional Intermediate Steps} contains details examples of several rewarded hacking cases, the SAE training process, and the distribution of feature activations found through specific layers of the SAE.
    \item \textbf{Appendix Additional Experiments and Full results} provides the full results and analyses of the experiment.
\end{itemize}

\section{Notations}
\label{sec_app:notation}
In this section, we briefly describe the symbols that we mainly use in this article. In table \ref{tab:notation}, we give the definitions of notation according to their role.
\begin{table*}
    \centering
    \renewcommand{\arraystretch}{1.2}
    \setlength{\tabcolsep}{1mm}
    \begin{tabular}{cc}
    \toprule
    \textbf{Notations} & \textbf{Definitions} \\
    \midrule
    \textit{Notations of Data} & \textit{Definitions of Data} \\
    \midrule
    $(x, y^*) \sim \mathcal{D}$ & A sample from the distribution of problems and their ground-truth answers. \\
    $x \in \mathcal{X}$ & An input problem instance. \\
    $y^*$ & The correct answer associated with input $x$. \\
    $s_t \in \mathcal{S}$ & The $t$-th reasoning step. \\
    $r^{(t)} = (s_1, s_2, ..., s_t)$ & Reasoning path up to step $t$. \\
    $\mathcal{H}_l$ & Hidden activations from layer $l$ used to train SAEs. \\
    \midrule
    \textit{Notations of Model} & \textit{Definitions of Model} \\
    \midrule
    $\pi_\theta$ & Policy model that generates next-step distributions. \\
    $R_\phi$ & Reward model that scores reasoning paths. \\
    $h_{l,k} \in \mathbb{R}^d$ & Hidden representation at token $k$ in layer $l$. \\
    $z_{l,k} \in \mathbb{R}^m$ & Sparse latent representation from the autoencoder. \\
    $W_e^{(l)}, W_d^{(l)}$ & Encoder and decoder weight matrices of the SAE. \\
    $f_i^{(l)}$ & Semantic feature basis vector $i$ in decoder. \\
    $\hat{h}_{l,k}$ & Reconstructed hidden representation. \\
    $\hat{R}_{CRA}(x)$ & CRA-adjusted reward score. \\
    \midrule
    \textit{Notations of Variables} & \textit{Definitions of Variables} \\
    \midrule
    $X$ & Reasoning path variable in the causal graph. \\
    $Y$ & PRM reward score. \\
    $Z$ & Spurious semantic feature acting as a confounder. \\
    $\tilde{z}_t^{(j)}$ & Intervened SAE latent vector at dimension $j$. \\
    $\mathcal{F}^\star$ & Set of identified confounding features. \\
    $t_j$ & $t$-statistic for feature $j$. \\
    $\mu_{1,j}, \mu_{0,j}$ & Mean activation of feature $j$ in positive/negative class. \\
    $\sigma^2_{1,j}, \sigma^2_{0,j}$ & Variance of feature $j$ in two classes. \\
    $P(Z=z)$ & Empirical prior distribution of feature $Z$. \\
    $E[Y \mid X, Z=z]$ & Conditional reward under fixed feature. \\
    \midrule
    \textit{Notations of Learning Objective} & \textit{Definitions of Learning Objective} \\
    \midrule
    $\mathbb{E}_{(x, y^*) \sim \mathcal{D}}[\mathbb{I}[\hat{y} = y^*]]$ & Expected accuracy over data distribution. \\
    $\mathcal{L}^{(l)}$ & SAE training loss combining reconstruction and sparsity. \\
    $E[Y \mid \text{do}(X=x)]$ & Causal effect of reasoning path $x$ on reward. \\
    $\arg\max_{(r,y)} R_\phi(x, r)$ & Selection of final output based on reward. \\
    \bottomrule
    \end{tabular}
    \caption{The definitions of notations used in our method.}
    \label{tab:notation}
\end{table*}

\section{Additional Related work}
\textbf{Sparse Autoencoders(SAEs)} have emerged as a powerful tool for mechanistic interpretability, particularly for decomposing neural network activations into interpretable, monosemantic features \cite{bricken2023monosemanticity}. Recent work has demonstrated that SAEs can successfully identify interpretable directions in language model activation spaces by resolving the superposition hypothesis \cite{cunningham2023sparseautoencodershighlyinterpretable}, with applications scaling to production-scale models like Claude 3 Sonnet that extract highly abstract, multilingual, and multimodal features \cite{templeton2024scaling}. 
However, principled evaluation of SAE features remains challenging due to the absence of ground-truth labels, leading to recent work on developing frameworks for comparing unsupervised feature dictionaries against supervised baselines \cite{makelov2024principledevaluationssparseautoencoders}. While prior SAE applications have primarily focused on understanding model capabilities and feature extraction, our work explores the application of SAEs to identify and intervene on specific failure modes like reward hacking in external reasoning systems.

\textbf{Causal Inference}
Causal inference is a framework that aims to uncover cause-and-effect relationships among variables, with its central goal being to understand how changes in one variable influence others. To support this, the Structural Causal Model (SCM) is widely adopted as a formal modeling approach, which represents causal dependencies through structural equations and a Directed Acyclic Graph (DAG). Based on this framework, the do-calculus is introduced to simulate interventions—explicitly setting a variable to a fixed value—in order to derive the resulting joint distribution, analogous to observing effects in randomized controlled trials \cite{pearl2009causality, peters2017elements}. Additionally, SCM provides the notion of $d$-separation, a graphical criterion for determining conditional independence between variables, which plays a crucial role in identifying confounding paths and ensuring the validity of causal inference \cite{spirtes2016causal}.

\section{Datasets}
\label{sec_app:dataset}
In this section, we provide a brief overview of the datasets used in our experiments. We utilize publicly available datasets. 

\begin{itemize}
    \item \textbf{MATH }\cite{hendrycksmath2021}. The MATH dataset is an open-source dataset created by Dan Hendrycks, Collin Burns, Saurav Kadavath, Akul Arora, Steven Basart, Eric Tang, Dawn Song, and Jacob Steinhardt, designed to measure mathematical problem-solving abilities. Released at NeurIPS 2021, the dataset contains a large number of math problems and is suitable for research in the fields of machine learning and artificial intelligence.
    \item \textbf{GSM8K} \cite{cobbe2021gsm8k}. The GSM8K dataset is a collection of grade school math problems released by OpenAI, containing 8.5K high-quality mathematical questions. These problems were created by human writers and are designed to support research in multi-step mathematical reasoning. The dataset is divided into 7.5K training problems and 1K test problems, each requiring 2 to 8 steps to solve, primarily using basic arithmetic operations (+ - / *).
\end{itemize}

\begin{figure*}[tbp]  
\centering
\begin{minipage}{0.48\linewidth}
   \includegraphics[width=\linewidth]{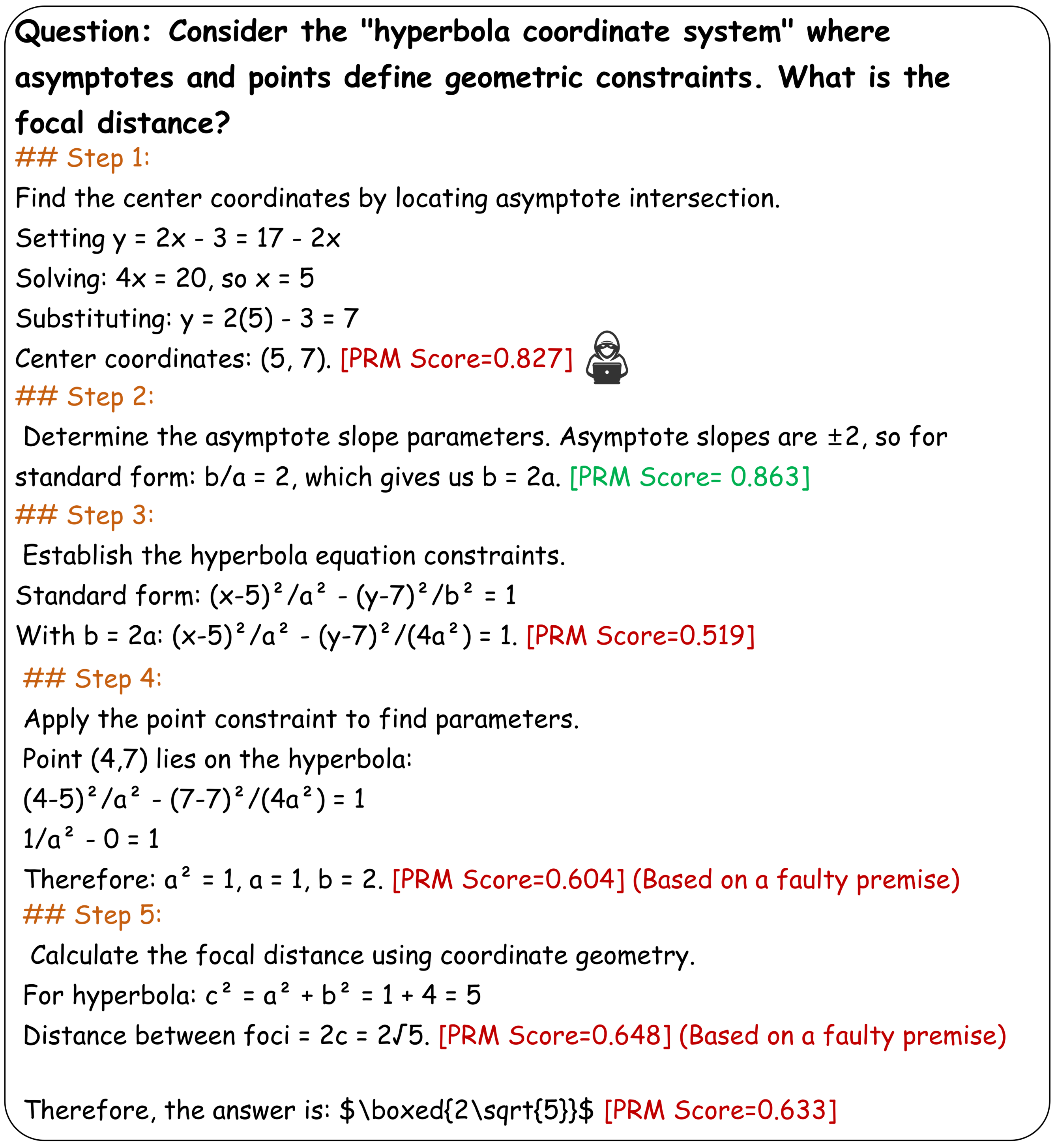}
   \caption*{(a) Hyperbola coordinate system problem}
\end{minipage}
\hfill
\begin{minipage}{0.48\linewidth}
   \includegraphics[width=\linewidth]{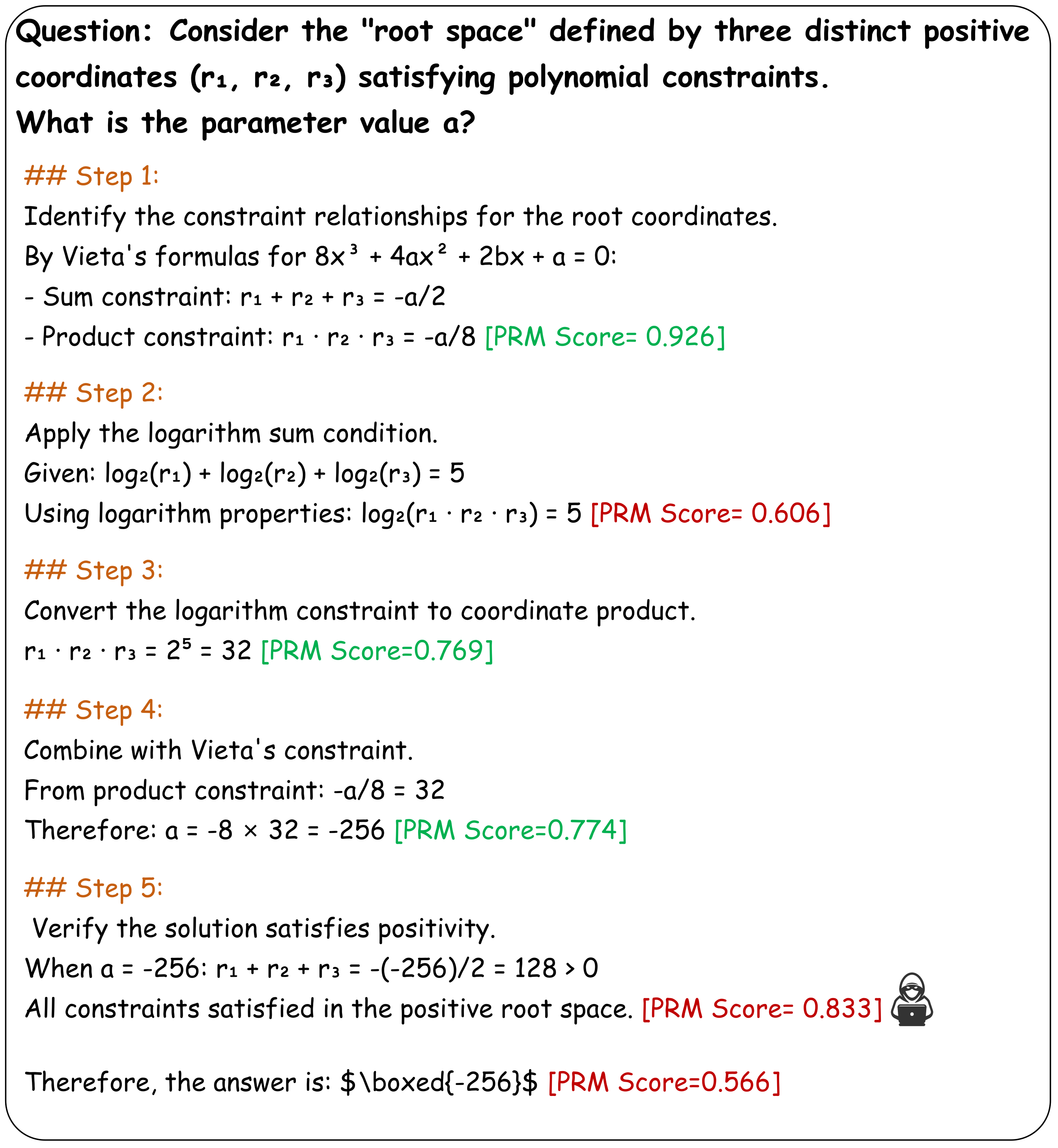}
   \caption*{(b) Polynomial root coordinate space problem}
\end{minipage}
\caption{More examples of reward hacking}
\label{figure:hacking_problem}
\end{figure*}

\section{LLMs}
\begin{itemize}
    \item \textbf{Qwen2.5-0.5B-Instruct }\cite{qwen2.5}. is an instruction-tuned language model from the Qwen2.5 series, with approximately 0.5 billion parameters. It is a causal language model based on the Transformer architecture, incorporating techniques such as Rotary Position Embedding (RoPE), SwiGLU activation, RMSNorm normalization, attention QKV biasing, and tied embeddings. The model undergoes both pretraining and post-training phases. It consists of 24 layers and uses Grouped Query Attention (GQA) with 14 query heads and 2 key-value heads. Qwen2.5-0.5B-Instruct supports a full context window of 32,768 tokens and can generate up to 8,192 tokens in a single pass. Among its 490 million parameters, 360 million are non-embedding parameters.
    \item \textbf{Qwen2.5-Math-7B-Instruct} \cite{yang2024qwen25mathtechnicalreportmathematical} is a 7-billion-parameter instruction-tuned large language model developed by Alibaba Cloud, specifically designed for mathematical reasoning tasks. As part of the Qwen2.5-Math series released in 2024, this model supports both English and Chinese math problem solving through Chain-of-Thought (CoT) and Tool-Integrated Reasoning (TIR). Compared to its predecessor Qwen2-Math, it shows significant improvements on mathematical benchmarks such as MATH. The model excels at symbolic computation, precise numerical reasoning, and algorithmic problem solving. It is fully open-sourced and compatible with the Hugging Face Transformers library (version $x \geq 0$ 4.37.0 is required for full functionality).
    \item \textbf{Llama-3.2-3B-Instruct} \cite{grattafiori2024llama3herdmodels} is a lightweight multilingual large language model released by Meta, employing an optimized Transformer architecture and supporting various tasks such as conversation, summarization, and rewriting. Through autoregressive generation, combined with Supervised Fine-Tuning (SFT) and Reinforcement Learning from Human Feedback (RLHF), this model achieves efficient performance and safety.
    \item \textbf{Math-Shepherd-PRM-7B} \cite{wang2023math} is an innovative, process-oriented mathematical reasoning reward model. It automatically constructs process supervision data, enabling step-by-step verification and reinforcement of LLMs mathematical reasoning abilities without requiring human annotation. The core innovation of this model lies in defining the quality of a reasoning step as its potential to derive the correct answer, and using "Completers" to automatically generate training data.
    \item \textbf{Qwen2.5-Math-PRM-7B} \cite{prmlessons} is a PRM released by Alibaba Cloud in 2025, designed for supervising mathematical reasoning processes. It identifies and corrects errors in the intermediate reasoning steps of large language models. This model is fine-tuned from Qwen2.5-Math-7B-Instruct, supports both English and Chinese, and is primarily used to evaluate the rationality of each reasoning step rather than generating text.
\end{itemize}

\begin{figure*}[tb]  
\centering
\includegraphics[width=1.0\linewidth]{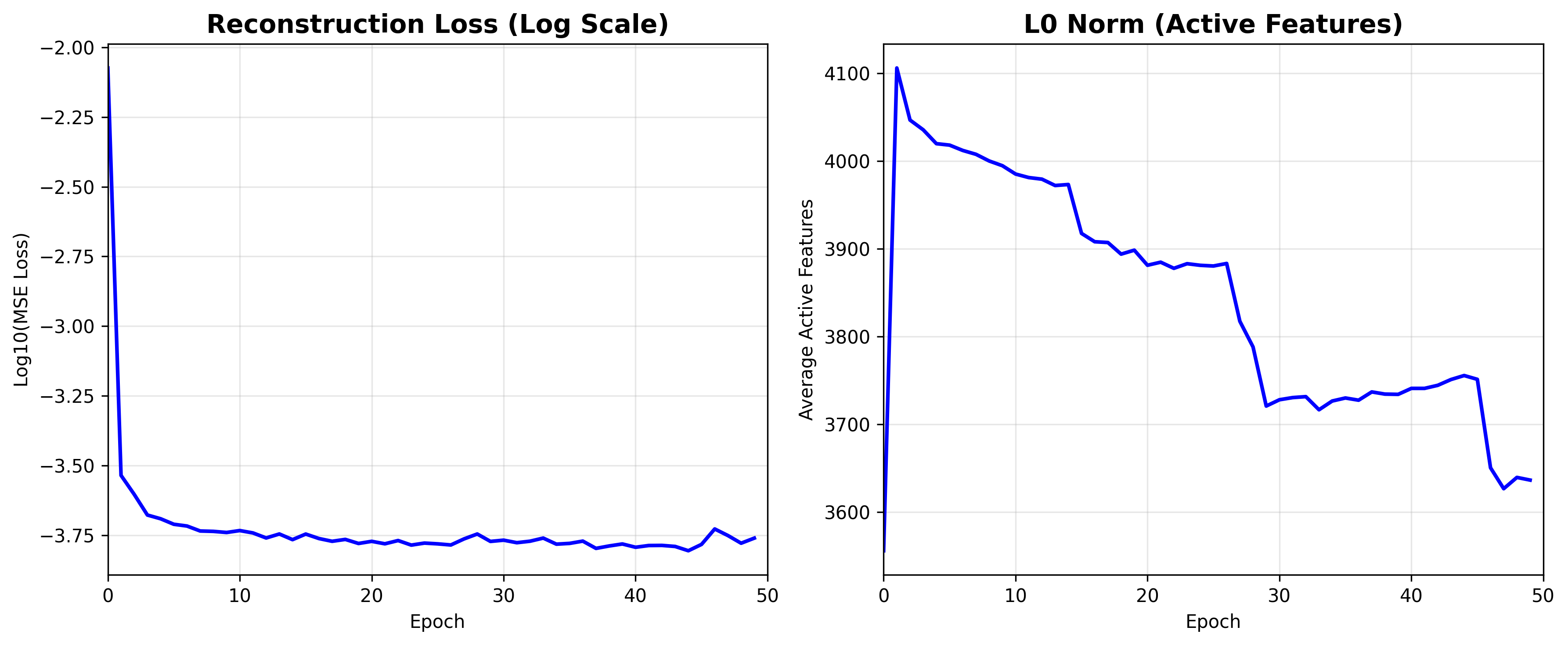}
\caption{SAE training process for the 16th layer Transformer Block of Math-Shepherd-PRM-7B}
\label{figure:SAE_training}
\end{figure*}

\section{Additional Intermediate Steps}
\subsection{More examples of reward hacking}
In this part, we supplement several examples of reward hacking phenomena. In fact, in our experiments, examples exhibiting reward hacking account for 49.5\% of all error cases.

In the hyperbola coordinate system problem (as shown in the first example in the figure~\ref{figure:hacking_problem}~(a)), reward hacking manifests as geometric concept error type. This problem determines the focal distance of a hyperbola through asymptote intersections and geometric constraints. The most severe reward hacking occurs in Step 1, which erroneously claims that "the given asymptotes are perpendicular to each other," while in reality, the two asymptotes have slopes of 2 and -2 respectively, with a slope product of $-4 \neq -1$, thus they are not perpendicular. This obviously incorrect geometric concept error received a high score of 0.827. Although Step 2 correctly determines the center coordinates $(5,7)$ through the asymptote intersection, subsequent Step 3 fails to properly utilize the asymptote slope information when establishing the hyperbola equation, resulting in a proposed equation that does not match the actual asymptotes. This case demonstrates how fundamental concept errors in geometric coordinate systems can receive inappropriately high rewards, thereby affecting the selection of the entire geometric constraint-solving strategy and ultimately deviating from the correct focal distance calculation path.

In the polynomial root coordinate space problem (as shown in the second example in the figure~\ref{figure:hacking_problem}~(b)), we identified a typical sign error type reward hacking phenomenon. This problem requires solving for parameter $a$ under constraint conditions, where three root coordinates $(r_1, r_2, r_3)$ must satisfy both Vieta's theorem constraints and logarithmic sum constraints. The key reward hacking occurs in Step 5, which perpetuates the fatal sign error from Step 2: incorrectly writing the root product relationship in Vieta's theorem as $r_1 \cdot r_2 \cdot r_3 = a/8$, when it should correctly be $r_1 \cdot r_2 \cdot r_3 = -a/8$ (missing the negative sign). This step, containing an obvious mathematical error, received a high score of 0.833, which is a typical reward hacking phenomenon. While the preceding steps (1, 3, 4) are correct in logarithmic property applications and coordinate transformations, this sign error directly causes the final answer to change from the correct $-256$ to the incorrect $256$, illustrating how a minor sign error can produce systematic bias in coordinate constraint systems.

\subsection{SAE Training Process}
We trained SAEs on each layer of the two PRM models selected in this paper to extract activation features. Figure~\ref{figure:SAE_training} illustrates the SAE training process for the 16th layer Transformer Block of Math-Shepherd-PRM-7B. The left plot shows a rapid drop in log-scale reconstruction loss, stabilizing after around 10 epochs, indicating effective representation learning. The right plot shows the average number of active features (L0 norm) decreasing from around 4100 to 3600, reflecting a successful sparsification process that balances reconstruction accuracy and feature compactness.

\begin{figure*}[tb]  
\centering
\includegraphics[width=1.0\linewidth]{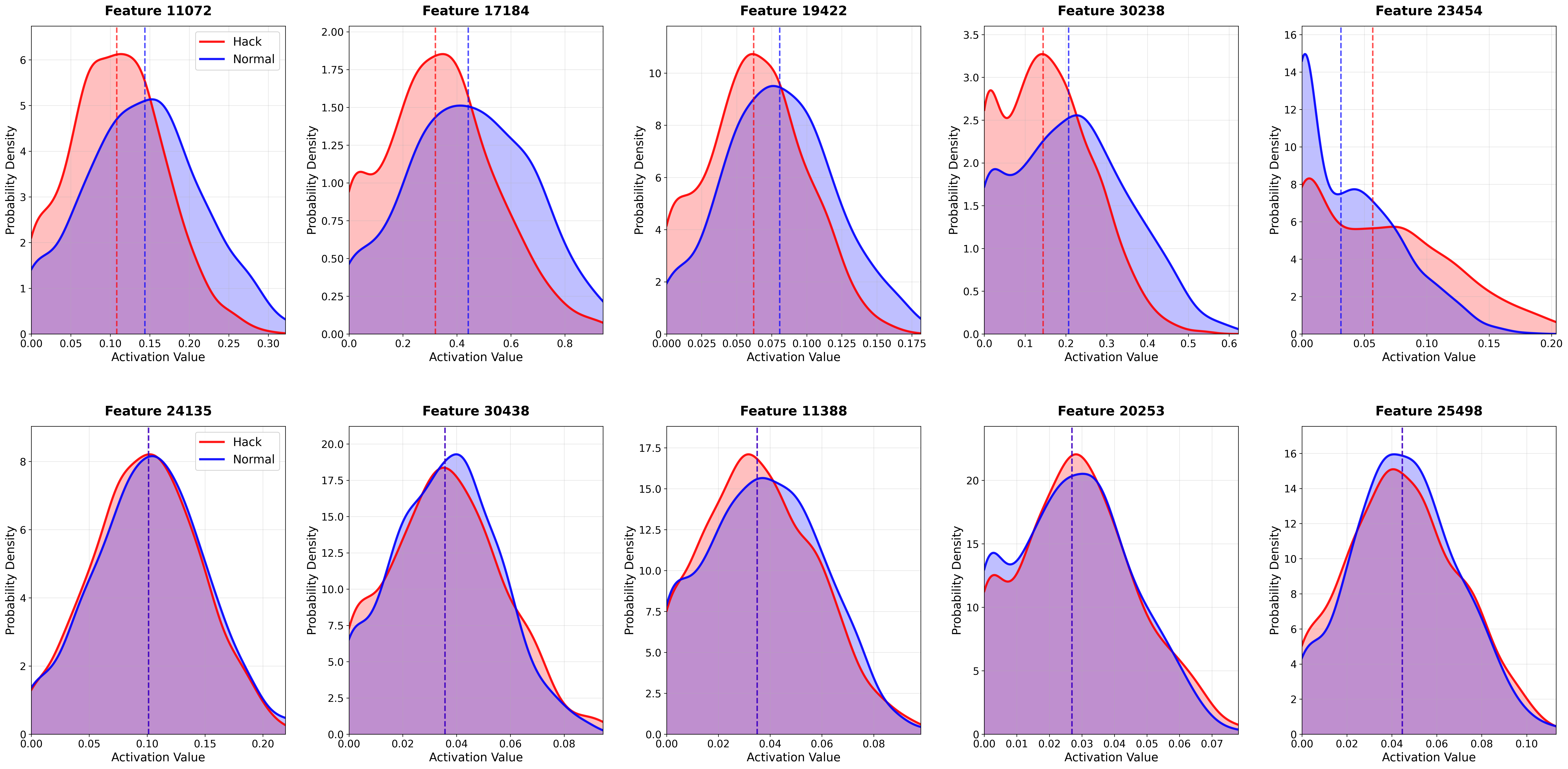}
\caption{Activation distributions of selected features ranked by $t$-statistics. The top row shows a subset of top-ranked features with the highest $t$-statistics; the bottom row shows features from the lower end of the ranking. Red and blue curves correspond to reward hacking and normal steps, respectively. Dashed lines indicate group-wise means. }
\label{figure:feature}
\end{figure*}

\subsection{Activation Distributions of Selected Features}
Figure~\ref{figure:feature} shows more activation value distributions for selected sparse features in the 16th transformer layer, chosen based on their $t$-statistics between reward hacking and normal samples. The top row illustrates features with high discriminative ability, where the activation means differ significantly between the two groups. In contrast, the bottom row shows features with low $t$-scores, exhibiting nearly identical activation distributions across reward hacking and normal cases.
\section{Additional Experiments and Full results}

\begin{figure*}[tb]  
\centering
\includegraphics[width=0.9\linewidth]{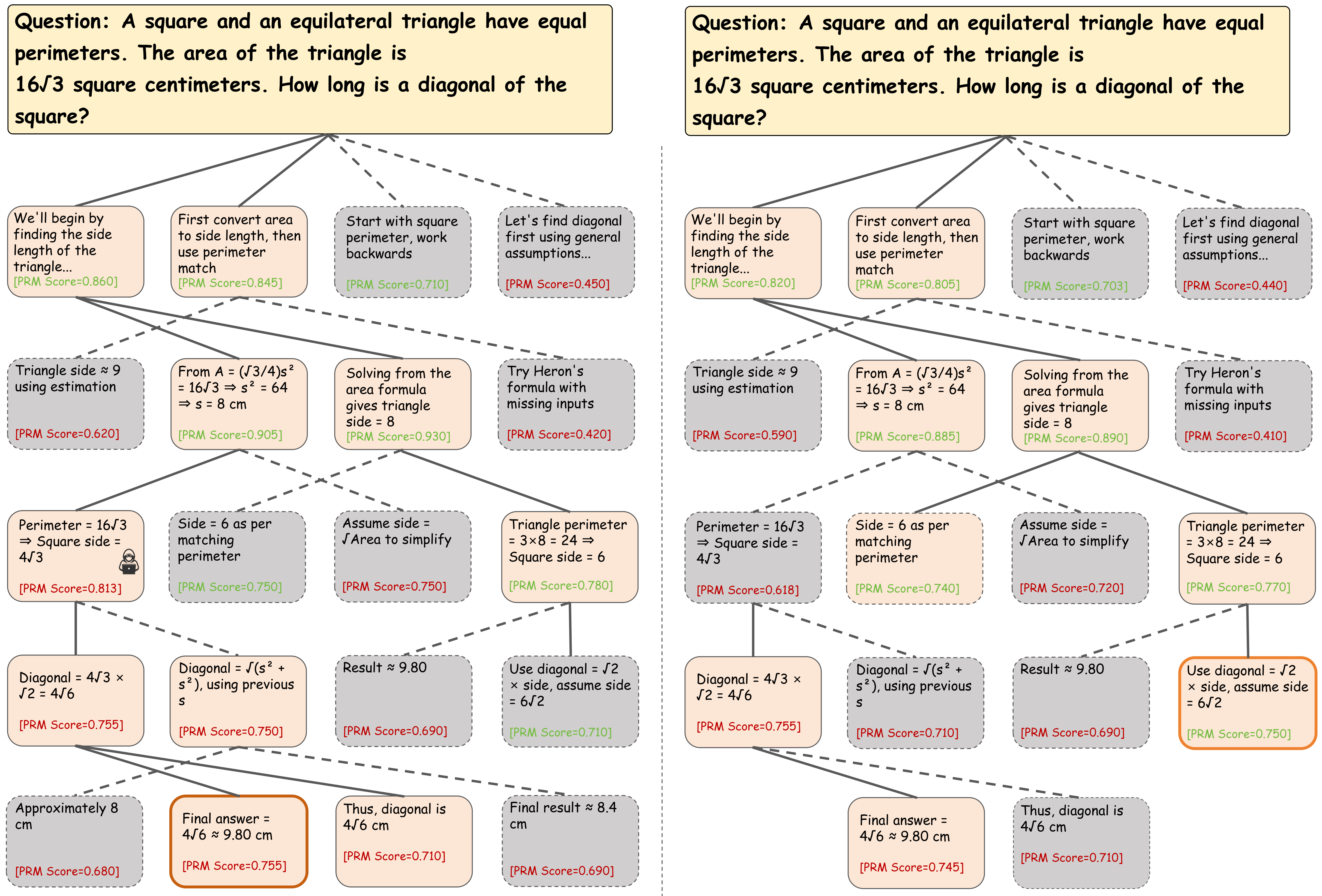}
\caption{ A comparison of Beam Search trees before (left) and after (right) intervention. Each node represents a reasoning step, with its corresponding PRM (Process Reward Model) score shown in the bottom-right corner. Steps enclosed by gray dashed borders indicate those selected in the final Beam Search path. Steps highlighted in red font with the hacker icon represent reward hacking cases—these steps contain clear logical errors but still receive high PRM scores, leading them to be mistakenly selected. Steps with only red font indicate incorrect reasoning that was appropriately assigned a low score.
}
\label{figure:change}
\end{figure*}

\subsection{The effect of intervening on reward hacking traits}
Figure~\ref{figure:change} presents a side-by-side comparison of beam search trees before (left) and after (right) applying feature-level interventions to the PRM. The reasoning task involves computing the diagonal of a square with the same perimeter as a triangle of known area.

On the left, without intervention, the model assigns a high PRM score (0.813) to an incorrect step that wrongly assumes the triangle’s area equals its perimeter. This step leads to a flawed path culminating in the wrong final answer ($4\sqrt{6}$). Despite earlier correct computations, this reward hacking step is favored due to its inflated score.

On the right, after applying our intervention, the same reward hacking step is penalized (PRM drops to 0.618), and the correct path—using the true triangle perimeter of 24—is promoted. As a result, the model selects the correct final answer ($6\sqrt{2}$), illustrating the effectiveness of intervention in disrupting harmful shortcuts and restoring alignment between PRM scores and reasoning correctness.

Overall, we observe a general decrease in PRM scores across nodes, particularly in reward-hacked candidates, while the reasoning structure remains intact. This confirms the intervention's ability to reduce the appeal of spurious patterns without breaking valid reasoning chains.

\begin{table*}[t]
\centering
\resizebox{\textwidth}{!}{%
\begin{tabular}{lcccccc}
\toprule
\textbf{Policy Model} & \multicolumn{2}{c}{\rule{0pt}{2.3ex}\textbf{Qwen2.5-0.5B-Instruct}\rule[-1.0ex]{0pt}{0pt}} 
& \multicolumn{2}{c}{\rule{0pt}{2.3ex}\textbf{Qwen-2.5-math-7B-Instruct}\rule[-1.0ex]{0pt}{0pt}} 
& \multicolumn{2}{c}{\rule{0pt}{2.3ex}\textbf{Llama-3.2-3B-Instruct}\rule[-1.0ex]{0pt}{0pt}} \\
\cmidrule{1-7}
\diagbox[dir=NW]{\textbf{Reward Model}}{\textbf{Dataset} \quad } & \textbf{MATH} & \textbf{GSM8K} & \textbf{MATH} & \textbf{GSM8K} & \textbf{MATH} & \textbf{GSM8K} \\
\cmidrule{1-7}
\rule{0pt}{11pt}Math-Shepherd-PRM-7B & 40.3 & 55.3 & 77.4 & 96.4 & 48.1 & 77.9 \\[2pt]
Qwen2.5-Math-PRM-7B & 46.4 & 61.2 & 77.9 & 96.7 & 53.6 & 80.3 \\[2pt]
\midrule
$\star$ Math-Shepherd-PRM-7B + \textbf{CRA} & 43.9 & 58.1 & 80.1 & 96.8 & 51.9 & 80.4 \\[2pt]
$\star$ Qwen2.5-Math-PRM-7B + \textbf{CRA} & \textbf{48.7} & \textbf{62.5} & \textbf{80.4} & \textbf{97.2} & \textbf{56.2} & \textbf{82.7} \\
\bottomrule
\end{tabular}%
}
\caption{Performance comparison using beam search (beam = 2) across different policy models. $\star$ represents our trained models.}
\label{tab:performance_2}
\end{table*}

\subsection{Supplementary Result}
Before setting the beam size to 4 in the main experiments, we also conducted evaluations with a smaller beam size of 2. As shown in Table~\ref{tab:performance_2}, CRA still brings consistent performance improvements under this constrained setting. On the MATH dataset, the average accuracy increases from 57.3\% to 60.2\%, with an average gain of 2.9 percentage points. On GSM8K, accuracy rises from 77.97\% to 79.62\%, yielding an average gain of 1.65 percentage points. The most notable improvements reach +3.80 on MATH and +2.80 on GSM8K, indicating that CRA remains effective even with limited candidate paths. These findings further support the robustness of CRA in mitigating reward hacking and enhancing external reasoning reliability.

\end{document}